\def\year{2022}\relax
\documentclass[letterpaper]{article} 
\usepackage{aaai22}  
\usepackage{amsfonts,amssymb} 
\usepackage{times}  
\usepackage{helvet}  
\usepackage{courier}  
\usepackage{amsmath}
\usepackage[hyphens]{url}  
\usepackage{graphicx} 
\usepackage{natbib}  
\usepackage{caption} 
\DeclareCaptionStyle{ruled}{labelfont=normalfont,labelsep=colon,strut=off} 
\usepackage{algorithm}
\usepackage{algorithmic}
\usepackage{verbatim}
\usepackage{bm}
\usepackage{multirow}
\usepackage{booktabs}
\usepackage{makecell}
\usepackage{graphics}
\usepackage{subfigure}

\newcommand{\etal}{{\emph{et al.}}}
\usepackage{newfloat}
\usepackage{listings}
\floatstyle{ruled}
\newfloat{listing}{tb}{lst}{}
\floatname{listing}{Listing}
\title{From Known to Unknown: Knowledge-guided Transformer for \\ Time-Series Sales Forecasting in Alibaba}
\author{
    Xinyuan Qi,
    Kai Hou,
    Tong Liu,
    Zhongzhong Yu,
    Sihao Hu,
    Wenwu Ou
}
\affiliations{
    Alibaba Group\\
    \{qishui.qxy,houkai.hk,sihao.hsh\}@alibaba-inc.com, \{yingmu,santong.oww\}@taobao.com, yuzzhongcs@163.com

}

\usepackage{bibentry}

\def\numx#1e#2{{#1}\mathrm{e}{#2}}

\begin{document}
\maketitle

\begin{abstract}
Time series forecasting (TSF) is fundamentally required in many real-world applications, such as electricity consumption planning and sales forecasting. In e-commerce, accurate time-series sales forecasting (TSSF) can significantly increase economic benefits. TSSF in e-commerce aims to predict future sales of millions of products. The trend and seasonality of products vary a lot, and the promotion activity heavily influences sales. Besides the above difficulties, we can know some future knowledge in advance except for the historical statistics. Such future knowledge may reflect the influence of the future promotion activity on current sales and help achieve better accuracy. However, most existing TSF methods only predict the future based on historical information. In this work, we make up for the omissions of future knowledge. Except for introducing future knowledge for prediction, we propose Aliformer based on the bidirectional Transformer, which can utilize the historical information, current factor, and future knowledge to predict future sales. Specifically, we design a knowledge-guided self-attention layer that uses known knowledge's consistency to guide the transmission of timing information. And the future-emphasized training strategy is proposed to make the model focus more on the utilization of future knowledge. Extensive experiments on four public benchmark datasets and one proposed large-scale industrial dataset from Tmall demonstrate that Aliformer can perform much better than state-of-the-art TSF methods. Aliformer has been deployed for goods selection on Tmall Industry Tablework, and the dataset will be released upon approval.

\end{abstract}
\section{Introduction}
Time series forecasting (TSF) plays a fundamental guiding role in many real-world applications, such as weather prediction\cite{karevan2020transductive}, financial investment\cite{alhnaity2020new}, 
and sales forecasting\cite{qi2019deep}. 
In e-commerce, accurate time-series sales forecasting (TSSF) supports the merchants to inventory stocks scientifically. It helps the platform optimize the turnover volume by assigning more exposure to the products with higher predicted sales value.
Taking the 2020 Tmall's 11.11 Global Shopping Festival as an example, the total turnover of 498.2 billion RMB reflects the long-term collaboration among the supply chain, industries, and e-commerce platforms behind this seemingly short period of promotion activity. 
These cooperation operations, including factory production, merchant stocking, and marketing strategy planning, cannot be achieved alone without the time-series sales forecasting algorithm.


Classical time series models \cite{brockwell2009time,box2015time,seeger2016bayesian,seeger2017approximate} make predictions based only on historical observations by analyzing the sequence's periodicity/seasonality and trend. Recent deep learning methods\cite{bai2018empirical} demonstrate effectiveness in both accuracy and generalization ability by encoding abundant features to represent the hidden status at each timestamp and enables a decoder to capture the long-range dependency of the hidden status sequence.
Regardless of the design differences, these two kinds of approaches all obey the principle of causality, i.e., the model should consider only the information before the timestamp $t$ when predicting the observation value on $t$. 

\begin{figure}[tbp!]
\centering
\includegraphics[scale=0.45]{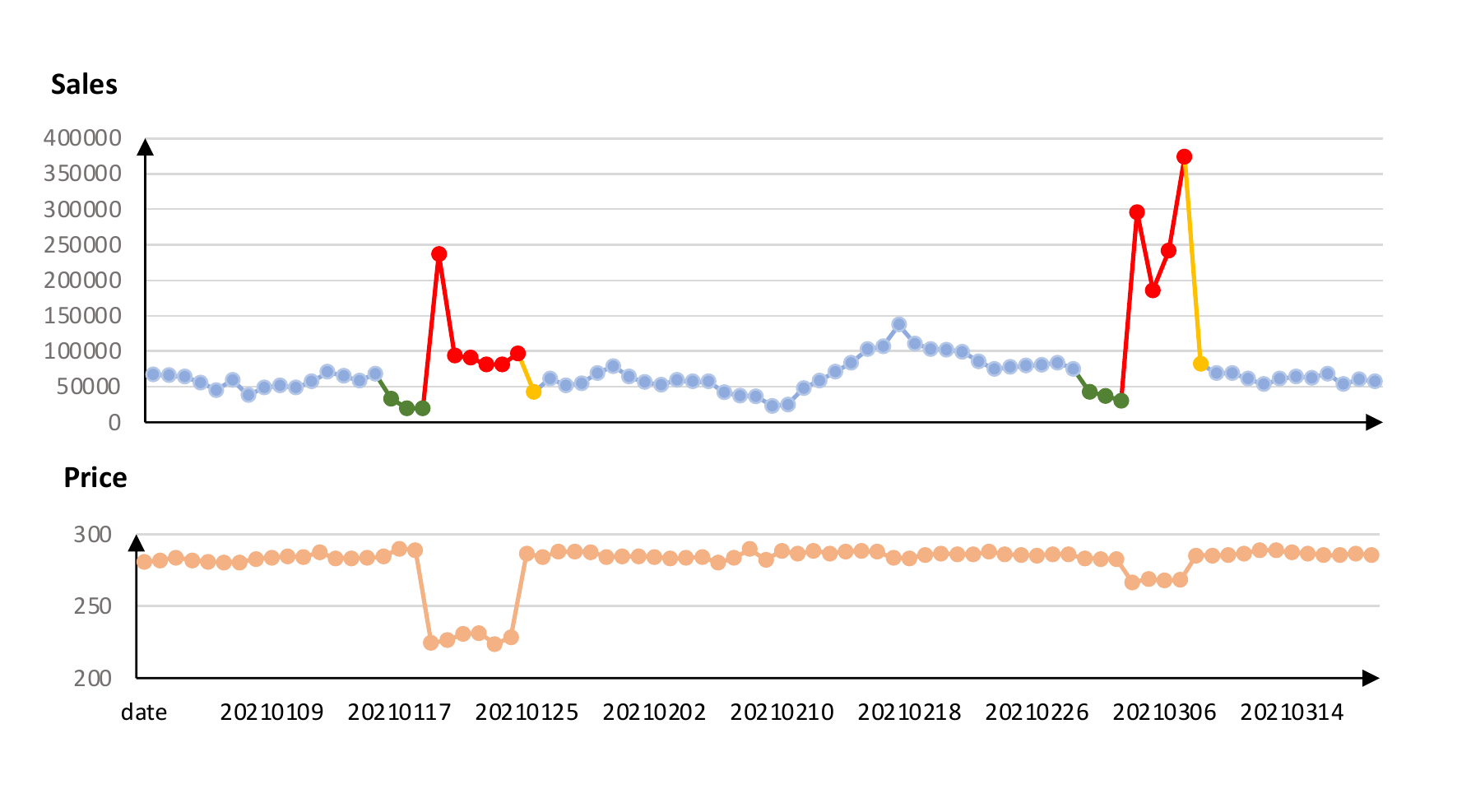}
\vspace{-15pt}
\caption{The time series of sales and price of an eye-shadow in Tmall experienced two promotion activities: one range from Jan 20 to Jan 25 and another from Mar 5 to Mar 8.}
\label{fig:fig1}
\end{figure}

However, some changeable information can be critical to the final results and even hurt the model's performance. To illustrate this point, we still use the sales forecasting task as an example: Figure~\ref{fig:fig1} shows the price of an eye-shadow over the past several months in Tmall, along with its corresponding sales volume. It is noted that the eye-shadow has experienced two relatively significant price reductions during promotion activities. Whenever the promotion approaches, the sales volume firstly decreases (doted in green) and then bursts when the activity begins (doted in red), suggesting that the future price information can largely influence the purchase behaviors. Consumers won't buy the product until the promotion starts.
And since existing methods can only predict the future through historical information, they will fail in predicting the sales volume during the promotion time caused by such changeable external factors, even though they can work well with regular time series.
Fortunately, a large part of future knowledge can be known ahead of time, especially for sales forecasting in e-commerce since sales value is heavily influenced by marketing operations like discount coupons setting, advertising investment, or participating in a live stream. All these operations can usually be known before the procedure is implemented because such operations generally require advance application and approval by the platform.



In this work, we consider the impact of the known future knowledge on TSSF and make up for shortcomings of existing methods that cannot leverage future knowledge in advance. We contribute the Knowledge-guided Transformer in Alibaba (\textit{\textbf{Aliformer}}) to fully make use of the knowledge information to guide the forecasting procedure.
Firstly, we define what information can be known in advance in our e-commerce scenario, which can be divided into two categories based on its sources: product- or platform-related, and will be described in more detail in the latter.

\begin{figure*}[h]
\centering
\includegraphics[scale=0.89]{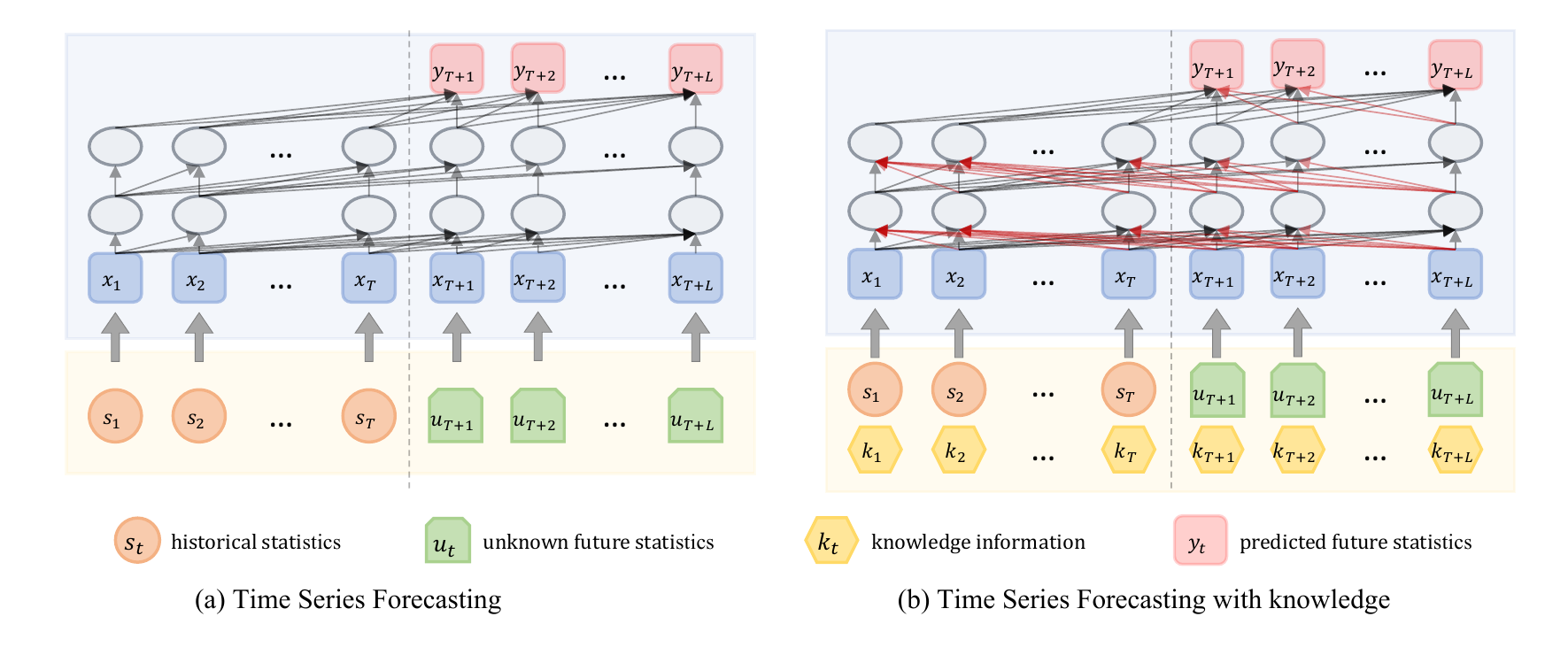}
\vspace{-10pt}
\caption{Illustration of the difference between time series forecasting task and time series forecasting with knowledge task.}
\label{fig:fig2}
\end{figure*}

Secondly, we describe our model given future knowledge above, which is mainly elaborated on the architecture of encoder of vanilla Transformer\cite{vaswani2017attention} by considering the recent prevalence and effectiveness of transformer-based methods in many sequence tackling applications. The significant discrepancy between these methods and our Aliformer is showed in Figure~\ref{fig:fig2}, which is that we allow leakage of future information by applying bidirectional self-attention on the input sequence with the known future information and anonymous information masked by a trainable token embedding.
However, directly applying bidirectional self-attention to the whole input sequence can lead to some unsatisfied results. The part of the series that is masked off can be seen as embeddings without any semantic information. Taking them as the key vectors to calculate the attention scores with other unmasked representations will bring some noise to the attention map. Even though our Aliformer has made up for the robustness of the masked token embedding with part of known future knowledge, the problem will still exist if we simply add knowledge embedding and masked token embedding together.
Therefore, we present the \textit{\textbf{AliAttention}} by adding a knowledge-guided branch to revise the attention map to minimize the impact of noise.
In general, to be consistent with the final task, we should mask off the last part of the sequence. But this may hinder the bidirectional learning ability of our approach because it will make the model rely on historical information other than future information.
Thus, we present the future-emphasized training strategy by adding span masking at the middle of the sequence to emphasize the importance of future knowledge. The hyper-parameter analysis further validates the effectiveness of this strategy.


Finally, we conduct extensive experiments on both e-commerce dataset and public time-series forecasting datasets. The evaluation shows that Aliformer sets the new state-of-the-art performance for time-series sales forecasting.
We also notice that there is a lack of real-world benchmark datasets for TSSF task. Therefore, we elaborately construct the Tmall Merchandise Sales (TMS) dataset, collected from the Tmall (the B2C e-commerce platform of Alibaba). TMS includes millions of time series sales data spanning over 200 days.
To our knowledge, it is the first public e-commerce dataset that tailored for time-series sales forecasting task. 
The dataset will be released upon approval.

In addition, we have implemented Aliformer into a prototype and deployed it on the ODPS system of Alibaba since May 1, 2021. We will select the top 1 million products out of billion products to participate in the promotion activities according to the predicted sales value to maximize the overall turnover for the platform. 
The evaluation of Aliformer on GMV coverage rate with \textbf{4.73} AP gain over the well-known Informer further illustrates its effectiveness.

\noindent \textit{\textbf{Contributions}} To summarize, the key contributions of this paper are:
\begin{itemize}
  \item  \textit{\textbf{Method}} We propose a knowledge-guided transformer (Aliformer) to fully make use of the future knowledge that largely affects sales value. Some strategies are proposed to enhance the ability of knowledge guidance. Extensive evaluations show Aliformer sets the new state-of-the-art performance for time-series sales forecasting.

  \item \textit{\textbf{Dataset}} A large benchmark dataset is collected from the Tmall platform to make up for the lack of an e-commerce dataset for the sales forecasting problem. The dataset will be released upon approval.

  \item \textit{\textbf{Application}} We deploy Aliformer in Alibaba, the largest e-commerce platform in China, and achieve significant performance improvements in the real-world application.
\end{itemize}


\section{Related Work}

\subsubsection{Time Series Forecasting} Various methods have been well developed in TSF. In the past years, a bunch of deep learning-based methods has been proposed to model the temporal dependencies for accurate TSF \cite{lai2018modeling, huang2019dsanet, chang2018memory}. 
LSTNet \cite{lai2018modeling} introduces the convolutional neural network (CNN) with a recurrent-skip structure to extract both the local dependencies and long-term trends. 
DeepAR \cite{salinas2020deepar} combines traditional autoregressive models and recurrent neural networks (RNNs) to model a probabilistic distribution for future time series.
Temporal convolution networks (TCNs) \cite{bai2018empirical, sen2019think} attempt to model the temporal causality with the causal convolution and dilated convolution. 
In addition, attention-based RNNs \cite{shih2019temporal, qin2017dual} incorporate the temporal attention to capturing the long-term dependencies for forecasting. 

Recently, the well-known Transformer \cite{vaswani2017attention} has achieved great success in language modeling \cite{devlin2018bert}, computer vision \cite{parmar2018image} \etal. Also, there are many Transformed-based methods for TSF \cite{zhou2021informer, li2019enhancing, wu2021autoformer}. LogTrans \cite{li2019enhancing} introduces causal convolutions into Transformer and proposes the efficient LogSparse attention.
Informer \cite{zhou2021informer} proposes another efficient ProbSparse self-attention for Long Sequence Time-Series Forecasting.
Autoformer \cite{wu2021autoformer} empowers the deep forecasting model with inherent progressive decomposition capacity through an Auto-Correlation mechanism. 

\subsubsection{Sales Forecasting}
As a typical application of TSF, sales forecasting plays a vital role in e-commerce platforms and has attracted significant interest. 
For general business forecasting tasks, \cite{taylor2018forecasting} proposes an analyst-in-the-loop framework named Prophet.
For extensive promotion sales forecasting in e-commerce, \cite{qi2019deep} propose a GRU-based algorithm 
to explicitly model the competitive relationship between the target product and its substitute products.
Another work presented by \cite{xin2019multi} fuses heterogeneous information
into a modified GRU cell to be aware of the status of the pre-sales stage before the promotion activities. 
For new product sales forecasting, \cite{ekambaram2020attention} utilize several attention-based multi-modal encoder-decoder models to leverage the sparsity of historical statistics of new products.

Yet all these approaches mainly focus on leveraging the historical statistics or enhanced multi-modal information to promote the accuracy, while ignoring the impact of some critical known future information on sales forecasting.
Unlike the above methods, we release some future information leakage yet do not break the principle of causality to capture some exciting correlation between sales value and some changeable factors. 

\section{Methodology}
Unsimilar to the TSF task, which forecasts the future observations based on the historical statistics, future sales in e-commerce are heavily influenced by the changeable external factors like the approaching promotion activity. Fortunately, we can know part of the external factors in advance. It's essential to predict sales with this future knowledge. In this section, we specify the definition of the research problem and explain the future knowledge, the vanilla self-attention layer for time series forecasting with knowledge. Then we present the proposed AliAttention layer, which treats the knowledge information as guidance. A future-emphasized training strategy is also proposed to make the model focus more on future knowledge. Finally, the panorama of the Aliformer is illustrated.

\subsection{Problem Formulation}
Given a sequence of historical statistics and knowledge information, the time series sale forecasting task aims to forecast the statistics in a future period. Let $n$ denotes a product, its historical statistics and knowledge information can be represented as a chronological sequence:
\begin{small}
$$ \mathbb{S} = \left\{s_1^{(n)},s_2^{(n)},...,s_T^{(n)}\right\} $$
$$ \mathbb{K} = \left\{k_1^{(n)},k_2^{(n)},...,k_{T}^{(n)}\right\} $$
\end{small}
where the statistics $s_t^{(n)}$ represents the historical statistics at $t$-th time (e.g. historical payed amount \etal), $k_t^{(n)}$ denotes the $t$-th time knowledge information (e.g. price at $t$-th time, which may vary at each time).

For time $t$, input $x_t^{(n)}$ can be represented with an embedding layer $\text{Emb}$:
\begin{small}
$$ x_t^{(n)} = \text{Emb}\left(\left\{s_t^{(n)},k_t^{(n)}\right\}\right)  \quad  {1 \le t \le T}$$
\end{small}
where $\text{Emb}$ means an FC layer for numerical features and a lookup table for id features, which map the features to $\mathbb{R}^{d_x}$. The sum of numerical and id features make up the input $x_t^n$.

Given the historical information of product $n$, the time series sale forecasting method predict the future sales can be formulated as:
\begin{small}
$$\left\{y_{T+1}^{(n)},...,y_{T+L}^{(n)}\right\} = f_h\left(x_1^{(n)},...,x^{(n)}_T\right) $$
\end{small}
$f_h$ means predicting the future based on historical information.

\subsection{Future Knowledge}
\label{sec:futureKnowledge}
Knowledge information can be any factor that potentially determines sales and can be known in advance, categorized into two types: product-related or platform-related.

Product-related knowledge information is intrinsic and describes the product itself, including price, advertising investment, whether in promotion activity~\etal.
Platform-related knowledge information is related to the promotion activity, such as the level, time, category of the activity. We can usually know the knowledge of a future period in advance because the promotion activity is scheduled, and products should follow certain rules in the campaign (e.g., give a certain discount). That is $k_{t}^{(n)} \left({T+1 \le t \le T+L}\right)$ can be known and:
\begin{small}
$$ \mathbb{K} = \left\{k_1^{(n)},k_2^{(n)},...,k_{T+L}^{(n)}\right\} $$
\end{small}
With the future knowledge involved, the input $x_t^{(n)}$ can be represented as:
\begin{small}
$$ x_t^{(n)} =\left\{
\begin{array}{lr}
\text{Emb}\left(\left\{s_t^{(n)},k_t^{(n)}\right\}\right)  \quad  {1 \le t \le T} \\
\text{Emb}\left(\left\{u_t^{(n)},k_t^{(n)}\right\}\right) \quad   {T+1 \le t \le T+L}
\end{array}
\right.
$$
\end{small}
the time series sale forecasting be formulated as:
\begin{small}
$$\left\{y_{T+1}^{(n)},...,y_{T+L}^{(n)}\right\} = f_b\left(x^{(n)}_1,...,x^{(n)}_{T+L}\right) $$
\end{small}
\noindent where $f_b$ denotes predicting the future based on both the historical information and future knowledge, $u_i^{(n)}$ means a default value or learnable token.

\begin{figure}[tbp!]
\centering
\includegraphics[scale=0.37]{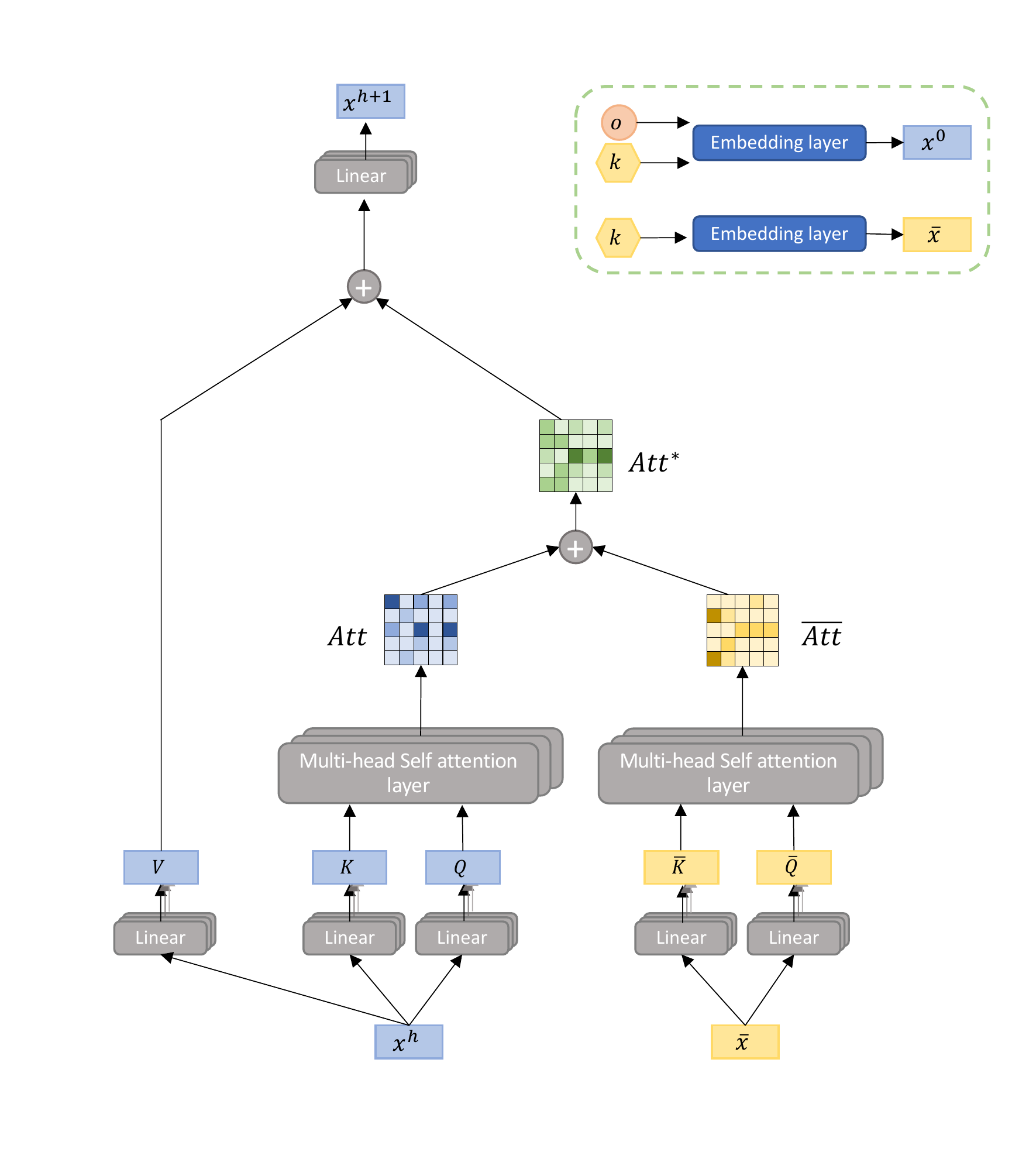}
\vspace{-15pt}
\caption{AliAttention layer. The knowledge-guided branch utilizes the pure knowledge information and outputs the knowledge-guided attention $\overline{\text{Att}}$ to revise the final attention map $\text{Att}^*$}
\label{fig:fig3}
\end{figure}

\subsection{Vanilla Self-Attention}
Bidirectional encoder from transformers is first introduced in BERT\cite{devlin2018bert}, which serves as a language representation model and focuses on learning word embedding. As shown in Figure~\ref{fig:fig2} (b), we introduce the bidirectional framework into the TSSF task. Statistics and knowledge are represented as vectors $x_t$ with the embedding technique. During training, the future statistics $s_t (t \in [T+1, T+L])$ are replaced with a learnable token $u_t (t \in [T+1, T+L])$. The bidirectional model will try to predict the future sales based on the historical statistics and knowledge with the vanilla self-attention (VSA):
$$
\text{Att}\left(i,j\right) = \frac{\left(x^h_iW_{Q}\right)\left(x^h_jW_{K}\right)^T}{\sqrt{d}} \label{att1}
$$
$$
x_i^{h+1} = \sum\nolimits_j\text{Softmax}(\text{Att}\left(i,j\right))\left(x^h_iW_{V}\right)W \label{z}
$$
where $x^h_i$ is $x^{(n)}$ in $h$-th VSA layer at time $i$, $d$ is a scale factor, $W_{Q},W_{K},W_{V},W$ represent linear layer to compute $Q,K,V$ in vanilla self-attention and $x^{h+1}_i$ respectively. With vanilla self-attention, information transmission can be bidirectional, and the current observations can be influenced by historical information, future knowledge, and current factors simultaneously. In such a framework, representations can be updated layer by layer with the self-attention mechanism:

$$ x^{h+1} = \rm{VSA}(x^h)$$
$$ x^1 = \left(\text{Emb}\left(\left\{s_1,k_1\right\}\right),...,\text{Emb}\left(\left\{u_{T+L},k_{T+L}\right\}\right)\right) $$

For the VSA is a positional invariant function, we add position embedding in $x_t$ to encode the position information explicitly to assist the time knowledge in $k_t$. From this perspective, position and time information can serve as knowledge in any time series dataset.


\subsection{AliAttention Layer}
The encoder of the Transformer can be viewed as stacked VSA layers. Statistics $s_t$ and knowledge $k_t$ is the component of input $x_t$. However, we don't know the future statistics $s_t (t \in [T+1,T+L])$ in fact. The tokenized training strategy will introduce noise while computing the attention map, and the learnable token may have the drawback of computing the attention map. Mixing the information from deterministic values and padded variables might make it unnecessarily difficult for the model to compute the relevance of each time.


\begin{figure}[tbp!]
\centering
\includegraphics[scale=1.2]{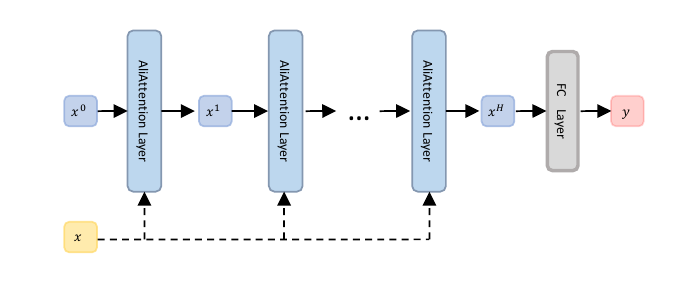}
\vspace{-20pt}
\caption{The overview of Aliformer. Original knowledge information is provided to each AliAttention layer.}
\label{fig:fig4}
\end{figure}

Therefore, we proposed the AliAttention layer, which utilizes the certainty of general knowledge and revises the attention map to minimize the impact of noise.
In detail, we modify the vanilla self-attention mechanism and add a knowledge-guided branch to revise the final attention map in the AliAttention layer. As shown in Figure~\ref{fig:fig3}, except for the vanilla self-attention, we introduce a branch for the pure knowledge information:
$$ \overline{x}_t^{(n)} = \text{Emb}\left(k_t^{(n)}\right)  \quad  {1 \le t \le T+L}$$

For the proposed AliAttention, it take both integrated information $x_t^{(n)}$ and knowledge information $\overline{x}_t^{(n)}$ as input. AliAttention compute attention values based on $x_t^{(n)}$ and $\overline{x}_t^{(n)}$ respectively. In practice, for the integrated input $x^{(n)} \in \mathbb{R}^{B \times T \times C}$ together with knowledge series $\overline{x}^{(n)} \in \mathbb{R}^{B \times T \times C}$, where $B$ is batch size, $T$ is the length of time series, $C$ is the size of embedding vectors. The AliAttention can be formalized as:

$$
\text{Att}\left(i,j\right) = \frac{\left(x^h_iW_{Q}^x\right)\left(x^h_jW_{K}^x\right)^T}{\sqrt{2d}} \label{att1}
$$

$$
\overline{\text{Att}}\left(i,j\right) = \frac{\left(\overline{x}_iW_{Q}^{\overline{x}}\right)\left(\overline{x}_jW_{K}^{\overline{x}}\right)^T}{\sqrt{2d}}\label{att2}
$$

$$
\text{Att}\left(i,j\right)^* = \text{Att}\left(i,j\right) + \overline{\text{Att}}\left(i,j\right)\label{att}
$$

\vspace{-6pt}
$$
x_i^{h+1} = \sum\nolimits_j\text{Softmax}(\text{Att}\left(i,j\right)^*)\left(x^h_iW_{V}^x\right)W \label{z}
$$
where $x^h_i$ and $\overline{x}_i$ are the representation in $h$-th AliAttention layer at time $i$ of $x^{(n)}$ and $\overline{x}^{(n)}$ respectively. $\overline{x}^{(n)}$ won't be update layer by layer.  $d$ is a scale factor, $W^x_{Q},W^x_{K},W^x_{V}$ represent linear layer to compute vanilla $Q,K,V$. $W^{\overline{x}}_{Q},W^{\overline{x}}_{K}$ are the linear layer to compute the $Q,K$ of the knowledge-guided branch.
Knowledge series $\overline{x}^{(n)}$ contains clean known knowledge without any noise (padded value and learnable token). Thus the knowledge-guided attention $\overline{\text{Att}}$ performs as reviser to the final attention map $\text{Att}^*$.

The panorama of Aliformer is presented in Figure~\ref{fig:fig4}. Each AliAttention layer takes the integrated information and the original knowledge information as the input. Representations with the same dimension are output and will be fed into the next layer. The integrated information at the first layer is embedded statistics and knowledge information.  
To guarantee the guidance of knowledge, identical original knowledge is explicitly provided for each AliAttention layer.

\subsection{Future-emphasized Training Strategy}
TSSF task focuses on predicting sales of a period future. In the training stage, a naive strategy is to train the model with future statistics replaced with the learnable token and predict the future sales. However, the learnable token may drawback the model's ability to utilize the future knowledge because the learnable token may be a bias for the future information and make the model relies more on the history while predicting. To lighten this phenomenon, we propose a future-emphasized training strategy. That is, in the training stage, we introduce span masking\cite{joshi2020spanbert} in addition to the naive strategy. Series trained in span masking try to predict a period of sales in the middle time, which urges the model to emphasize the future information. One series is trained in the naive and span masking strategies with the probability $p_1$ and $p_2$, respectively. The effectiveness of our future-emphasized training strategy is confirmed in the experiment.

\section{Experiment}
\subsection{Datasets}
We conducted extensive experiments to evaluate the proposed model on four public benchmark datasets (ETTh\footnote[1]{\url{https://github.com/zhouhaoyi/ETDataset}}, ETTm\footnotemark[1], ECL\footnote[2]{\url{https://archive.ics.uci.edu/ml/datasets/ElectricityLoadDiagrams20112014}}, Kaggle-M5\footnote[3]{\url{https://www.kaggle.com/c/m5-forecasting-accuracy/data}}) and one real-world dataset (Tmall Merchandise Sales Dataset -- \textit{TMS}).
More details about the datasets are illustrated in the Appendix. Here we only take a brief introduction of \textit{TMS} dataset.

\subsubsection{Tmall Merchandise Sales Dataset}
We collect 1.2 million samples of product sales from the Tmall platform, the B2C e-commerce platform of Alibaba. The time-series sales data span from Sep 13, 2020, to Apr 15, 2021 (215 days). Each sample contains 86-dimensional features, including price, category, brand, date and statistical information of product, seller, and category.
The five most relevant features to products are listed in the following, which are also the targets in experiments:
\begin{itemize}

    \item[-] item page view (\textit{\textbf{ipv}}): The daily views count of the product details page.
    
    \item[-] unique visitor (\textit{\textbf{ipv\_uv}}): The unique daily visitors' count of the product details page.
    
    \item[-] gross merchandise volume (\textit{\textbf{gmv}}): The daily deal amount of the product.
    
    \item[-] count (\textit{\textbf{ord}}): The order count of the product sold per day.
    
    \item[-] buyer (\textit{\textbf{byr}}): The buyer count of the product sold per day.
    
\end{itemize}

The scale of \textit{TMS} is much larger and more complex than the current standard public datasets. In the actual scenario of sales forecast, we could get richer future information, such as product prices, marketing activities status in \textit{TMS}. We conduct mainly analysis of our method on \textit{TMS} in the following sections.


    
    
    
    
    

\subsection{Experimental Details}
We select several recently time series forecasting methods as our baselines: (1) LSTNet \cite{lai2018modeling} (2) LSTMa \cite{bahdanau2015neural} (3) LogTrans \cite{li2019enhancing} (4) Informer \cite{zhou2021informer}
More experimental details are put in the Appendix to avoid trivializing description.

\subsection{Results and Analysis}
\subsubsection{Comparison with the state-of-the-art}

\begin{table*}[t!]
  \centering
  \caption{Time series forecasting results of all methods on five datasets. The number in brackets means the target value's dimensions. The best results are highlighted.}
    \begin{tabular}{c|c|cc|cc|cc|cc|cc}
    \toprule
    \multicolumn{2}{c|}{Methods} & \multicolumn{2}{c|}{Aliformer} & \multicolumn{2}{c|}{Informer} & \multicolumn{2}{c|}{LogTrans} & \multicolumn{2}{c|}{LSTMa} & \multicolumn{2}{c}{LSTNet} \\
    \midrule
    \multicolumn{2}{c|}{Metric} & MSE   & MAE   & MSE   & MAE   & MSE   & MAE   & MSE   & MAE   & MSE   & MAE \\
    \toprule
    TMS(5) & 15    & \textbf{0.154 } & \textbf{0.229 } & 0.321  & 0.353  & 0.327  & 0.368  & 0.313  & 0.354  & 0.283  & 0.336  \\
    \midrule
    Kaggle-M5(1) & 28    & \textbf{0.526 } & \textbf{0.555 } & 0.552  & 0.568  & 0.544  & 0.570  & 0.533  & 0.556  & 0.528  & 0.561  \\
    \midrule
    \midrule
    \multirow{4}{*}{ETTh(7)} & 48    & \textbf{0.767 } & \textbf{0.694 } & 1.575  & 1.086  & 1.952  & 1.122  & 1.805  & 1.094  & 3.629  & 1.697  \\
          & 168   & \textbf{1.480 } & \textbf{0.957 } & 3.166  & 1.480  & 3.693  & 1.642  & 4.449  & 1.559  & 3.218  & 1.984  \\
          & 336 & \textbf{1.604 } & \textbf{1.056 } & 2.933  & 1.446  & 4.173  & 1.902  & 3.706  & 1.500  & 4.606  & 2.027  \\
          & 720   & \textbf{2.455 } & \textbf{1.383 } & 2.903  & 1.410  & 3.276  & 1.485  & 3.961  & 1.659  & 4.972  & 3.847  \\
    \midrule
    \multirow{4}{*}{ETTm(7)} & 48    & \textbf{0.309 } & \textbf{0.371 } & 0.413  & 0.444  & 0.424  & 0.515  & 0.647  & 0.607  & 1.672  & 1.072  \\
          & 96 & \textbf{0.324 } & \textbf{0.376 } & 0.574  & 0.538  & 0.651  & 0.694  & 0.788  & 0.677  & 2.340  & 1.351  \\
          & 288   & \textbf{0.430 } & \textbf{0.452 } & 0.823  & 0.687  & 1.140  & 1.154  & 1.190  & 0.908  & 0.980  & 1.814  \\
          & 672   & \textbf{0.581 } & \textbf{0.551 } & 1.134  & 0.868  & 1.588  & 1.370  & 2.107  & 1.306  & 1.824  & 2.758  \\
    \midrule
    \multirow{4}{*}{ECL(321)} & 48    & \textbf{0.304 } & \textbf{0.369 } & 0.387  & 0.428  & 0.399  & 0.455  & 0.486  & 0.572  & 0.443  & 0.446  \\
          & 168   & \textbf{0.330 } & \textbf{0.396 } & 0.393  & 0.435  & 0.394  & 0.443  & 0.574  & 0.602  & 0.381  & 0.420  \\
          & 336 & \textbf{0.322 } & \textbf{0.386 } & 0.389  & 0.424  & 0.380  & 0.432  & 0.886  & 0.795  & 0.419  & 0.477  \\
          & 720   & \textbf{0.396 } & \textbf{0.408 } & 0.424  & 0.454  & 0.427  & 0.466  & 1.676  & 1.095  & 0.556  & 0.565  \\
    \bottomrule
    \end{tabular}%
  \label{overall-results}%
\end{table*}%

The evaluation results of all the methods on five datasets are summarized in Table~\ref{overall-results}. As can be seen, the Transformer-based methods, the Aliformer, Informer, and LogTrans, show better results than LSTMa and LSTNet.

The proposed method-Aliformer achieves consistent state-of-the-art performance on all five datasets and all prediction lengths. Especially, the Aliformer greatly outperforms other comparison methods on the real-world product sales dataset - \textit{TMS}. Our method gets a MSE improvement of 52\% (0.321 $\rightarrow$ 0.154) for Informer, 53\% (0.327 $\rightarrow$ 0.154) for LogTrans. The great improvement reveals that the proposed knowledge-guided AliAttention layer and future-emphasized training strategy can fully use this future knowledge for better predictions. The greatly gain on \textit{TMS} dataset and relatively poorer gain on other datasets(whose knowledge information only include position and time information) also reflect that future knowledge is vital for future forecasting accurately.

\subsubsection{Ablation Analysis}
To verify the effectiveness of future knowledge and AliAttention mechanism, we conduct ablation experiments on \textit{TMS} dataset for each target value, i.e., \textit{ipv}, \textit{ipv\_uv}, \textit{amt}, \textit{cnt}, and \textit{byr}. The results of each target value and all target values are reported in Table~\ref{ablation-results}.

\vspace{5pt}
\begin{table}[htbp]
  \centering
  \footnotesize
  \renewcommand\arraystretch{0.9}
  \caption{Ablation results of the proposed Aliformer.}
  \vspace{-5pt}
    \begin{tabular}{c|cc|cc|cc}
    \toprule
    \multirow{2}[4]{*}{TMS} & \multicolumn{2}{c|}{wo/ future} & \multicolumn{2}{c|}{wo/AliAttention} & \multicolumn{2}{c}{Aliformer (full)} \\
\cmidrule{2-7}  & MSE   & MAE   & MSE   & MAE   & MSE   & MAE \\
    \midrule
    $ipv$   & 0.065  & 0.185  & 0.059  & 0.176  & 0.059  & 0.176  \\
    $ipv\_uv$ & 0.055  & 0.168  & 0.049  & 0.159  & 0.049  & 0.159  \\
    $gmv$   & 0.639  & 0.487  & 0.494  & 0.413  & 0.456  & 0.391  \\
    $ord$   & 0.130  & 0.245  & 0.110  & 0.219  & 0.105  & 0.213  \\
    $byr$   & 0.124  & 0.238  & 0.104  & 0.212  & 0.100  & 0.206  \\
    \midrule
    $Avg$ & 0.203  & 0.265  & 0.163  & 0.236  & \makecell[c]{\textbf{0.154}} & \makecell[c]{\textbf{0.229}} \\
    \bottomrule
    \end{tabular}%
  \label{ablation-results}%
\end{table}%
\vspace{5pt}

\begin{itemize}
  \item  \textbf{Effect of future knowledge} In this study, we modify the proposed Aliformer to predict without future knowledge, which predicts only based on historical information. As shown in Table~\ref{ablation-results}, this ablation model "wo/ future" achieved a worse prediction with 0.203/0.265 for MSE/MAE. The significant reduction verifies that future knowledge is essential for future sales prediction. 

  \item \textbf{Effect of AliAttention layer} The knowledge-guided AliAttention layer is a core component in our Aliformer, which efficiently utilizes the consistency of general knowledge in future forecasting. Results in Table~\ref{ablation-results} between the ablation model "wo/ AliAttention" and the full model show that this knowledge-guided AliAttention layer leverages the future-known knowledge with more comprehensive depictions for better prediction.
\end{itemize}

\vspace{5pt}
\begin{figure}[hbp]
\centering
\subfigure[AliAttention layer numbers.]{
\centering
\includegraphics[scale=0.31]{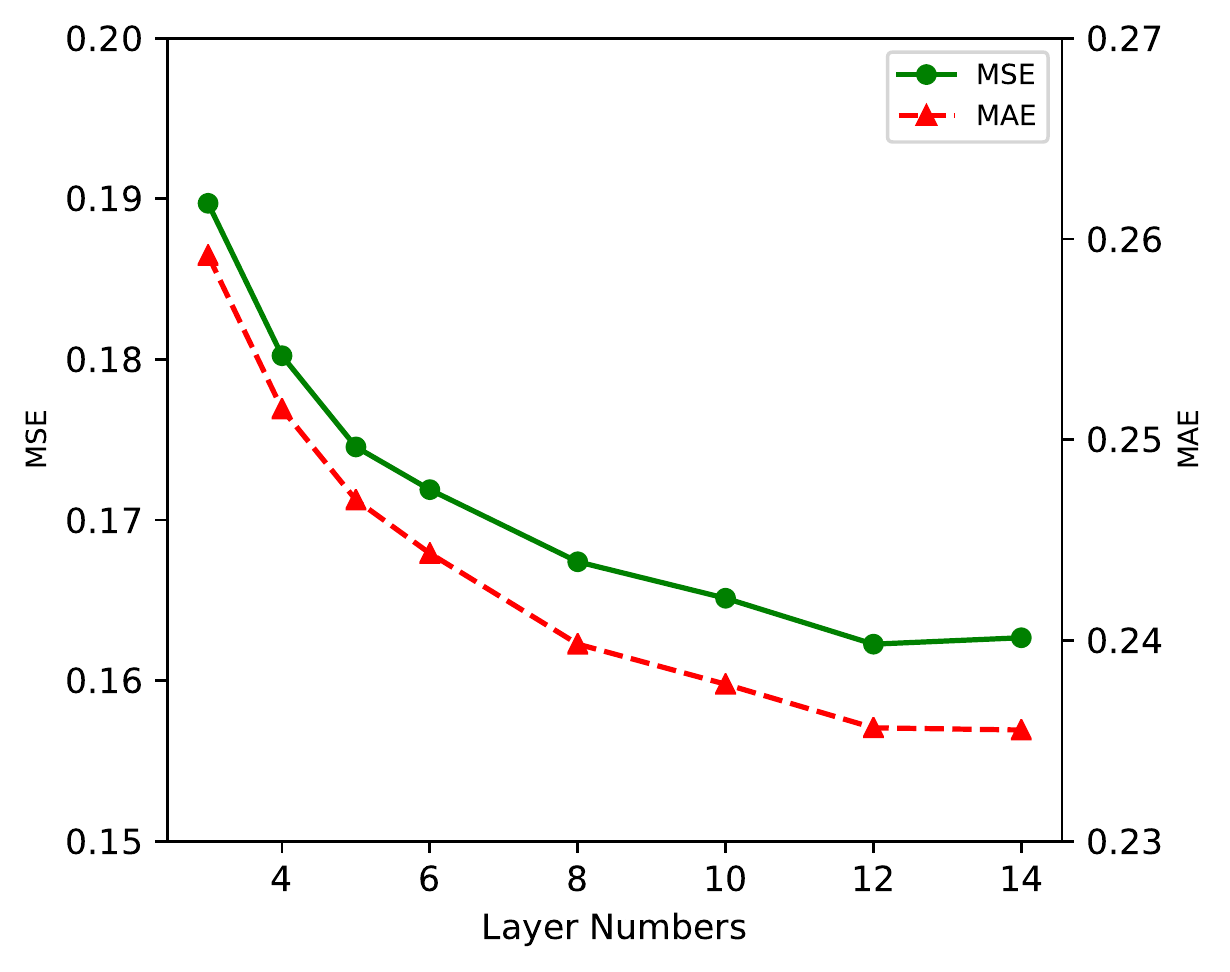}
}
\subfigure[Span masking probability.]{
\centering
\includegraphics[scale=0.31]{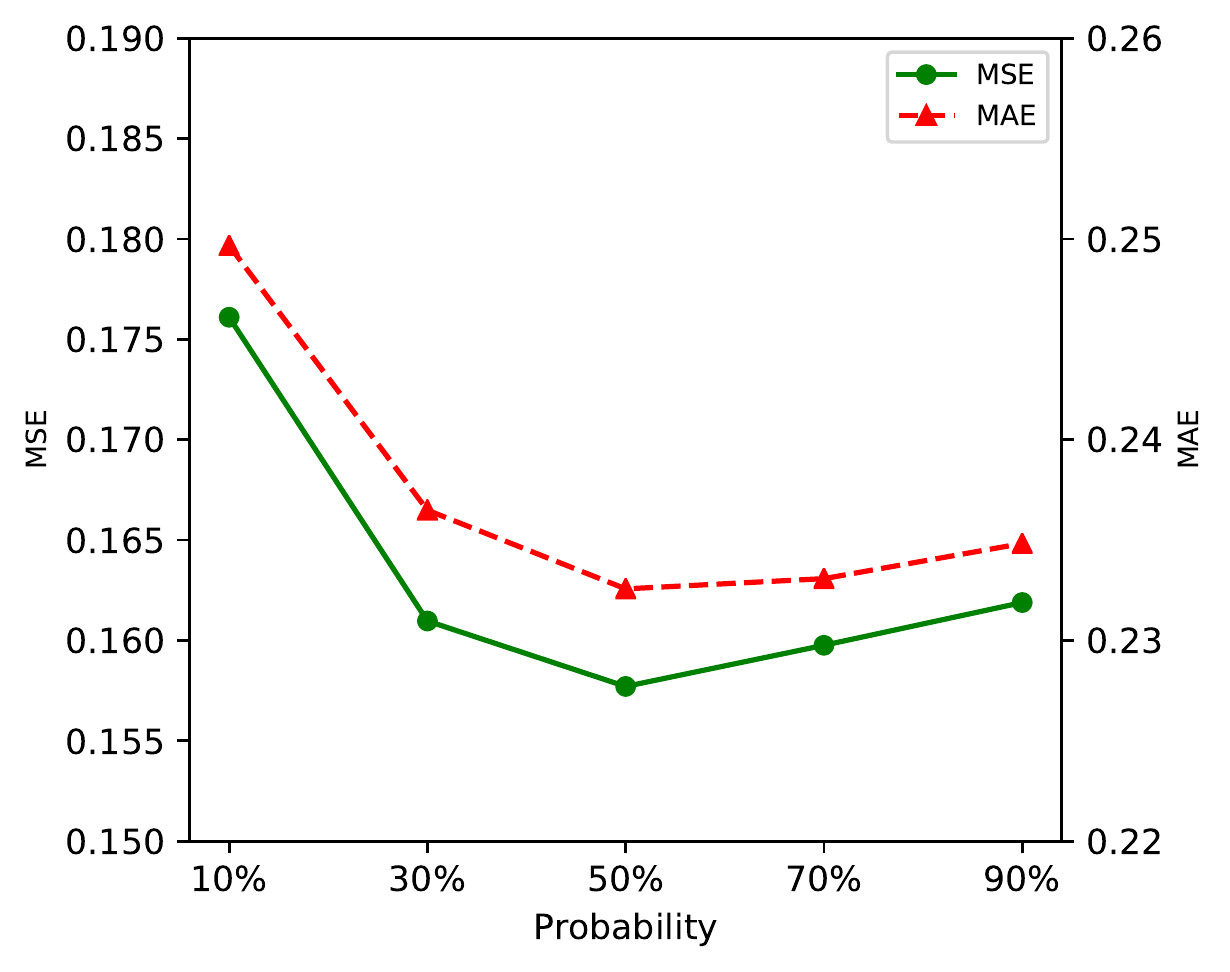}
}%
\vspace{-10pt}
\caption{The parameter sensitivity of two components in the proposed Aliformer.}
\label{fig:fig5}
\end{figure}



\subsubsection{Parameter Sensitivity}
We provide the sensitivity analysis of the Aliformer model on \textit{TMS} dataset for two important hyperparameters. \textit{\textbf{AliAttention Layer Numbers:}} In the Figure~\ref{fig:fig5} (a), we could find that the Aliformer model achieves better prediction results with more AliAttention layers. However, the improvement of increasing layers is minimal when the model contains 12 AliAttention layers. Thus, we set the number of AliAttention layers equal to 12 on \textit{TMS} dataset. \textit{\textbf{Span Masking Probability $p_2$:}} As we can see in the Figure~\ref{fig:fig5} (b), increasing this probability $p_2$ brings lower MSE and MAE, but further increasing degrades the performances. Because when the probability arrives at some great value, the model can leverage the future knowledge but ignores to predict future sales.
There is no significant performance change during $50\%-90\%$. We set the span masking probability $p_2$ as 50\% in practice. The experiments for parameter sensitivity express the capacity and robustness of the Aliformer model.

\begin{figure*}[tbp]
\centering
\subfigure[Forecasting comparison results.]{
\centering
\includegraphics[scale=0.41]{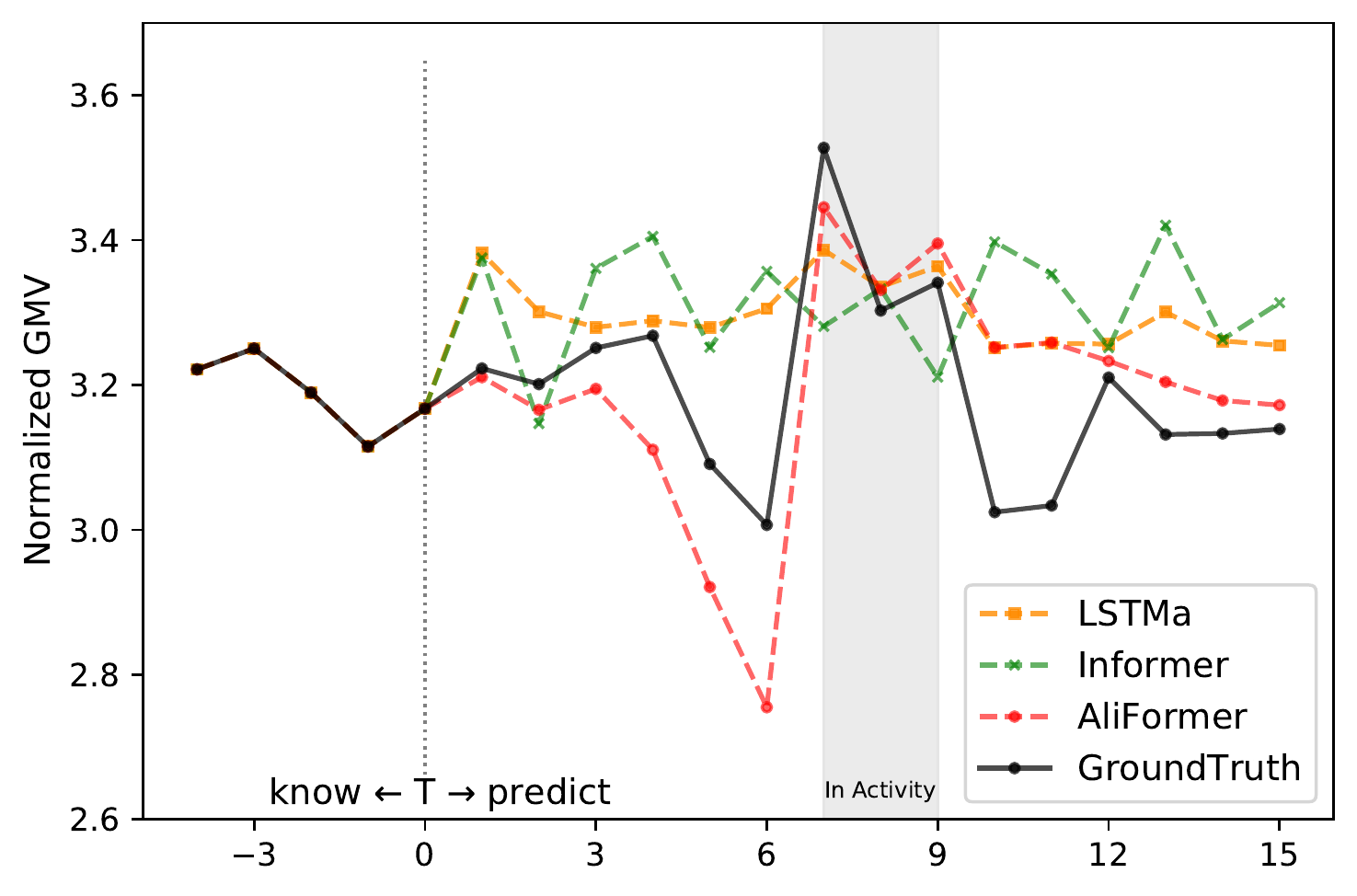}
}
\hspace{15pt}
\subfigure[Forecasting results with varying sales price.]{
\centering
\includegraphics[scale=0.41]{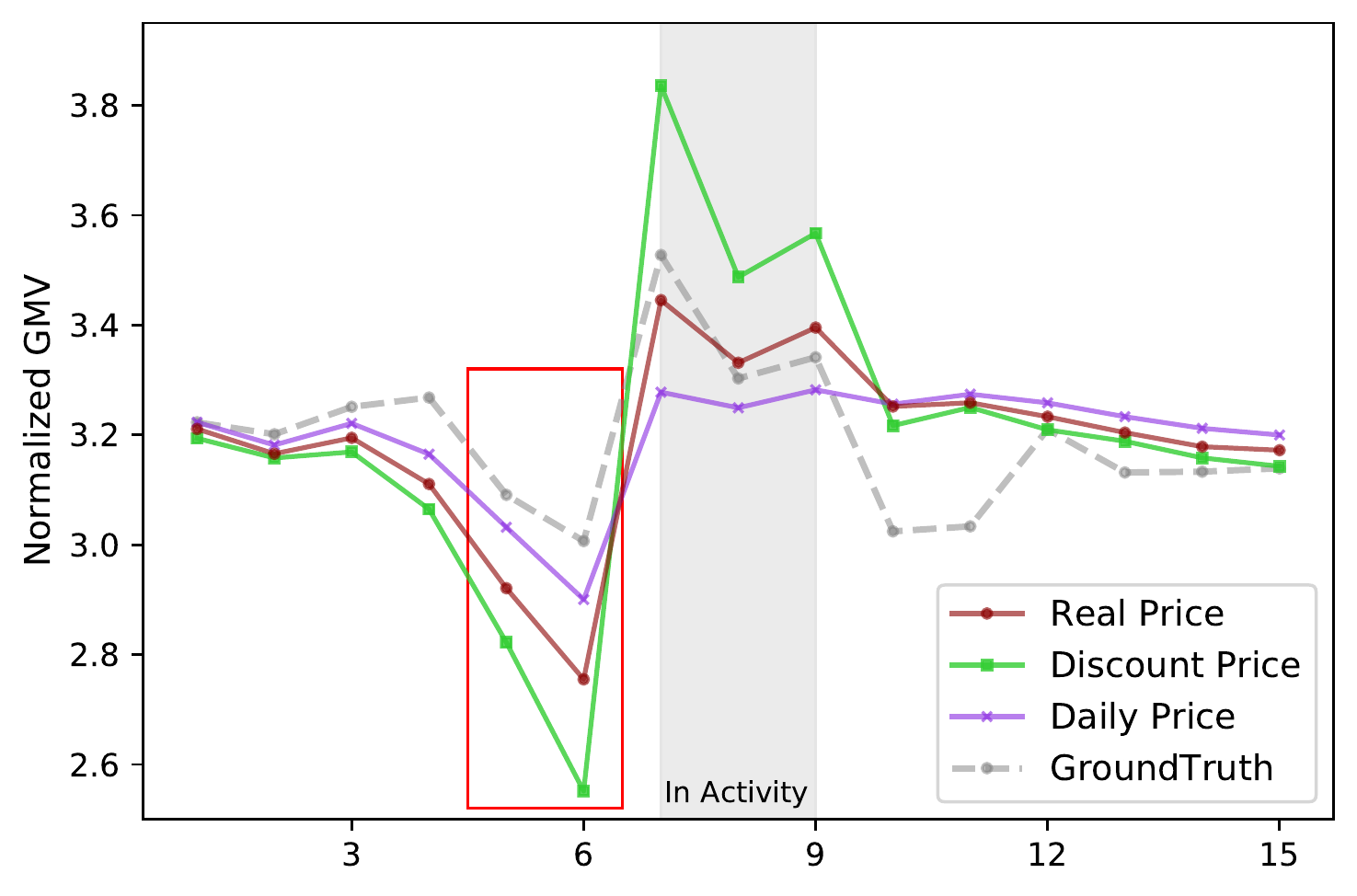}
}%
\vspace{-5pt}
\caption{Visualization results for case study on \textit{TMS}.}
\label{fig:fig6}
\end{figure*}
\vspace{5pt}


\subsubsection{Visualization analysis for case study}
We provide the visualization results for case study on \textit{TMS} dataset, including the forecasting comparison results and forecasting resluts of the proposed Aliformer varying the target product's sales price. In Figure~\ref{fig:fig6} (a), we show the future 15-days forecasts of the LSTMa, Informer and the proposed Aliformer. As we can see, the Aliformer achieves better prediction than other comparison methods. When predicting the future 15-days target value, the Aliformer could catch the sudden alteration before the marketing activity and perform good forecasting with the help of both the historical temporal patterns and future knowledge.

In Figure~\ref{fig:fig6} (b), we study the effect of the product sales price on product's sales \textit{gmv}. We perform this analysis by varying one product's sales price during a promotion activity. We show the prediction results when we set the sales price as the real price, the daily price (the averaged price of the former seven days), and a discount price (20\% off). The results show that a discount price would increase the sales performance during the promotion activity but further inhibit the performance before this activity. This phenomenon conforms to a general view in the real world that customers tend to buy products at a lower sales price and delay their purchase behavior until the promotion campaigns start. Such ability can also advise sellers to set prices in the marketing activity for their desire sales.

\subsubsection{More explanations on AliAttention mechanism}
\begin{figure}[h]
\centering
\subfigure[PDF of attention weights.]{
\centering
\includegraphics[scale=0.25]{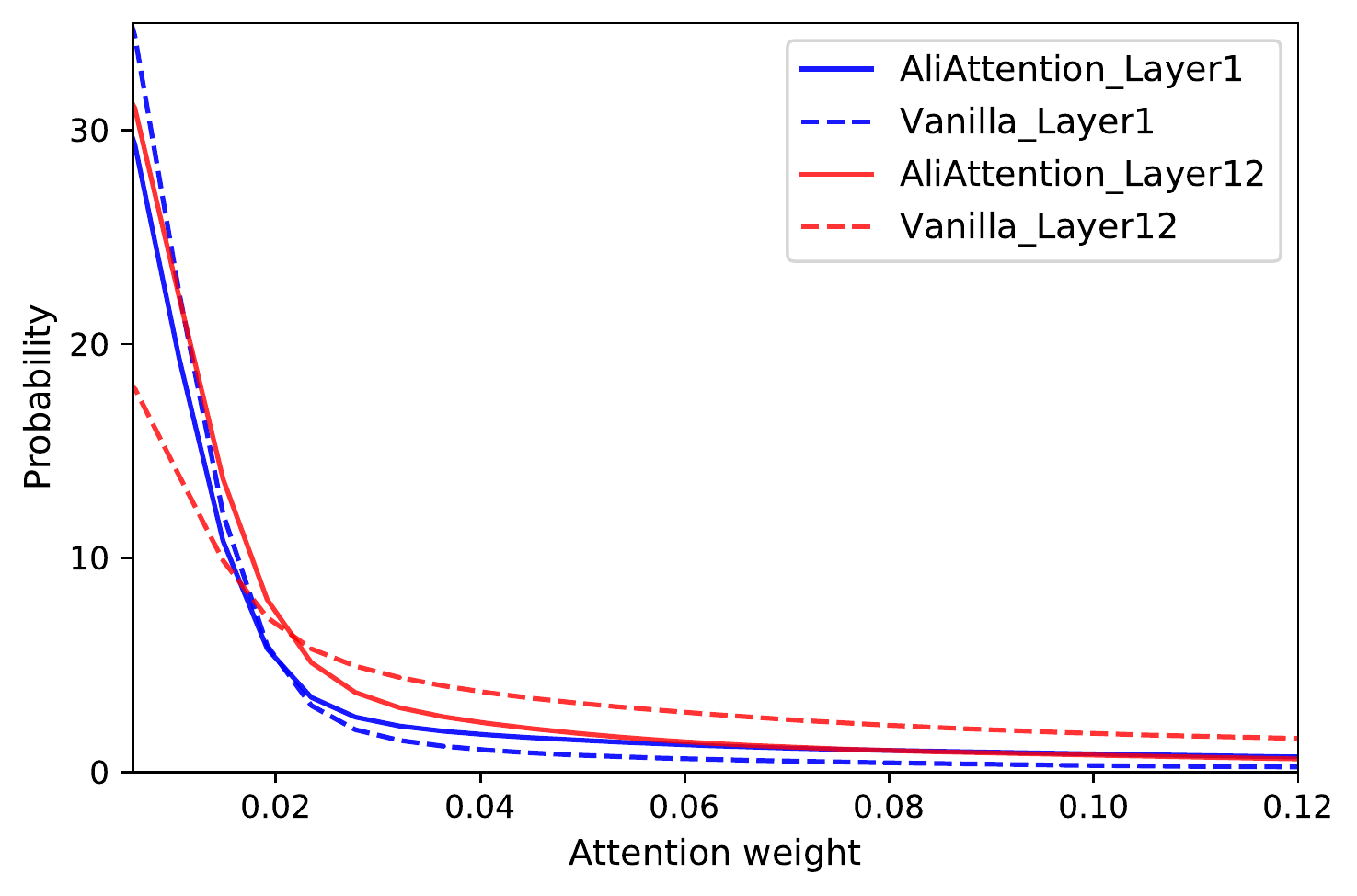}
}
\subfigure[Weight's proportion.]{
\centering
\includegraphics[scale=0.25]{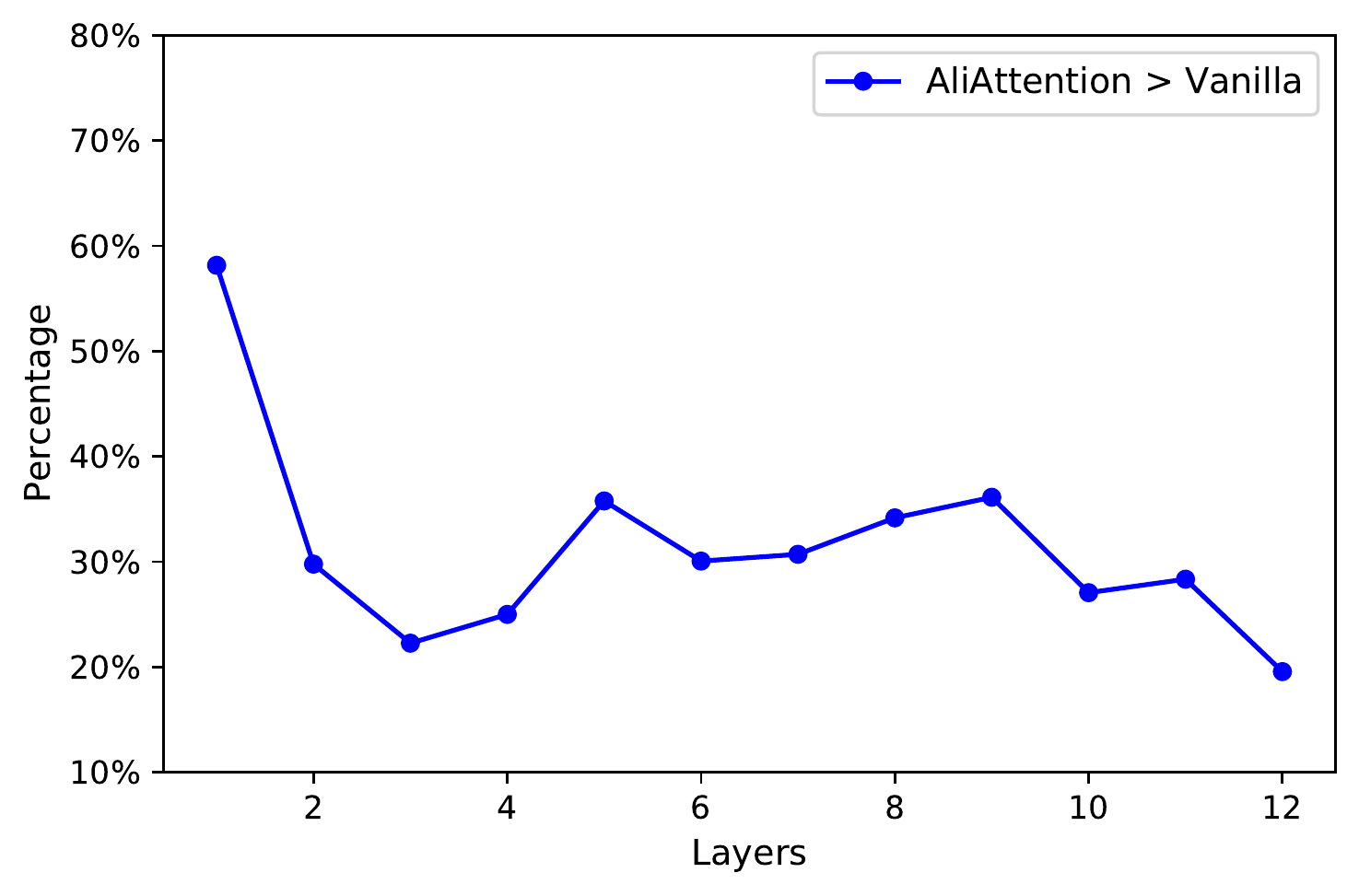}
}%
\vspace{-5pt}
\caption{Visualization results for the AliAttention's two major components: the vanilla attention $\text{Att}$ and the knowledge aligned attention $\overline{\text{Att}}$.}
\label{fig:fig7}
\end{figure}
\vspace{5pt}

Furthermore, we provide a visualization analysis for the proposed AliAttention mechanism to study how its two components, the vanilla attention $\text{Att}$ and the knowledge-guided attention  $\overline{\text{Att}}$ work for forecasting. In Figure~\ref{fig:fig7} (a), we show the distributions of $\text{Att}$'s and $\overline{\text{Att}}$'s attention weights in the bottom layer (Layer\_1) and the top layer (Layer\_12). Additionally, we compute the proportion of $\overline{\text{Att}}$'s weights $>$ $\text{Att}$'s weights in different stacked layers, which is exhibited in Figure~\ref{fig:fig7} (b). As is shown, we could find the knowledge-guided attention $\overline{\text{Att}}$ plays a greater role on bottom layers, whereas a smaller role on top layers (more greater attention value in bottom layer and less in top layer). This indicates that vanilla attention captures sufficient information under the guidance of general knowledge with the stacked layers increasing.

\section{Deployment in Alibaba}
To verify the effectiveness of the proposed method in a real-world scenario, Aliformer has been deployed since May 1, 2021, to use goods selection for extensive promotion activities in Tmall. Specifically, we will sort billions of items according to the predicted sales value for the next 15 days based on the past 200 historical statistics and extra knowledge. The top 1 million products will then be selected to participate in the platform's promotion activities to obtain additional user exposure and click to maximize the overall GMV of the platform. In practice, Informer has been chosen as our baseline since it significantly improves our original tree-based algorithm. We conduct the comparison during the past Tmall 618 shopping festival during Jun 1, 2021 to Jun 20, 2021, by pre-selecting the candidates sorted by Aliformer and Informer. The evaluation shows that the overall sales volume of products selected by Aliformer can coverage the 74.96\% of overall GMV, with an \textbf{4.73} absolute percentage gain over Informer (70.23\% of overall GMV), suggesting that Aliformer can bring tremendous profit benefit for the e-commerce platform.


\section{Conclusion}
In this work, we present a knowledge-guided transformer (Aliformer) for TSSF task. The proposed method can make full use of future knowledge and significantly improve sales forecasting accuracy in e-commerce. Extensive experiments demonstrate its effectiveness. We also tailor the TMS dataset to make up for the lack of e-commerce benchmark datasets for TSSF problem.
The deployment of Aliformer further achieves significant performance improvements in the real-world application.

\newpage

\noindent \textbf{\LARGE Appendices}

\section{Datasets Description}
Here we provide a detailed introduction to four public datasets used in our experiment and the TMS dataset collected from the real-world e-commerce application.

\subsection{Public Benchmark Datasets}
\begin{itemize}
    \item \textit{Electricity Transformer Temperature (ETT)} \footnote{\url{https://github.com/zhouhaoyi/ETDataset}} datasets contain the load and oil temperature of electricity transformers from one county of China. Two separate datasets as \textit{ETTh} for 1-hour-level and \textit{ETTm} for 15-minute-level are collected between July 2016 and July 2018 \cite{zhou2021informer}. The training, validation, and test set are split in chronological order by the ratio of 6:2:2.
    
    \item \textit{Electricity Consuming Load (ECL)} \footnote{\url{https://archive.ics.uci.edu/ml/datasets/ElectricityLoadDiagrams20112014}} dataset records the hourly electricity consumption (Kwh) of 321 clients from 2012 to 2014. The train, validation, and test set is split as the ratio of 7:1:2.
    
    \item \textit{Kaggle-M5} \footnote{\url{https://www.kaggle.com/c/m5-forecasting-accuracy/data}} dataset, generously made available by Walmart in Kaggle M5 competition, involves the unit sales of various products sold in the USA. This dataset contains the past unit sales as well as calendar-related features.
    The Kaggle-M5 involves the unit sales of 3049 products sold across ten stores. Thus, we convert the dataset into 30490 series of products. Due to the unavailability of \textit{Kaggle-M5}'s test set, we train models on the first 1913 days and validate them on the following 28 days.
\end{itemize}

\begin{figure}[t!]
\centering
\subfigure[Sales distribution of \textit{TMS} dataset.]{
\centering
\includegraphics[scale=0.4]{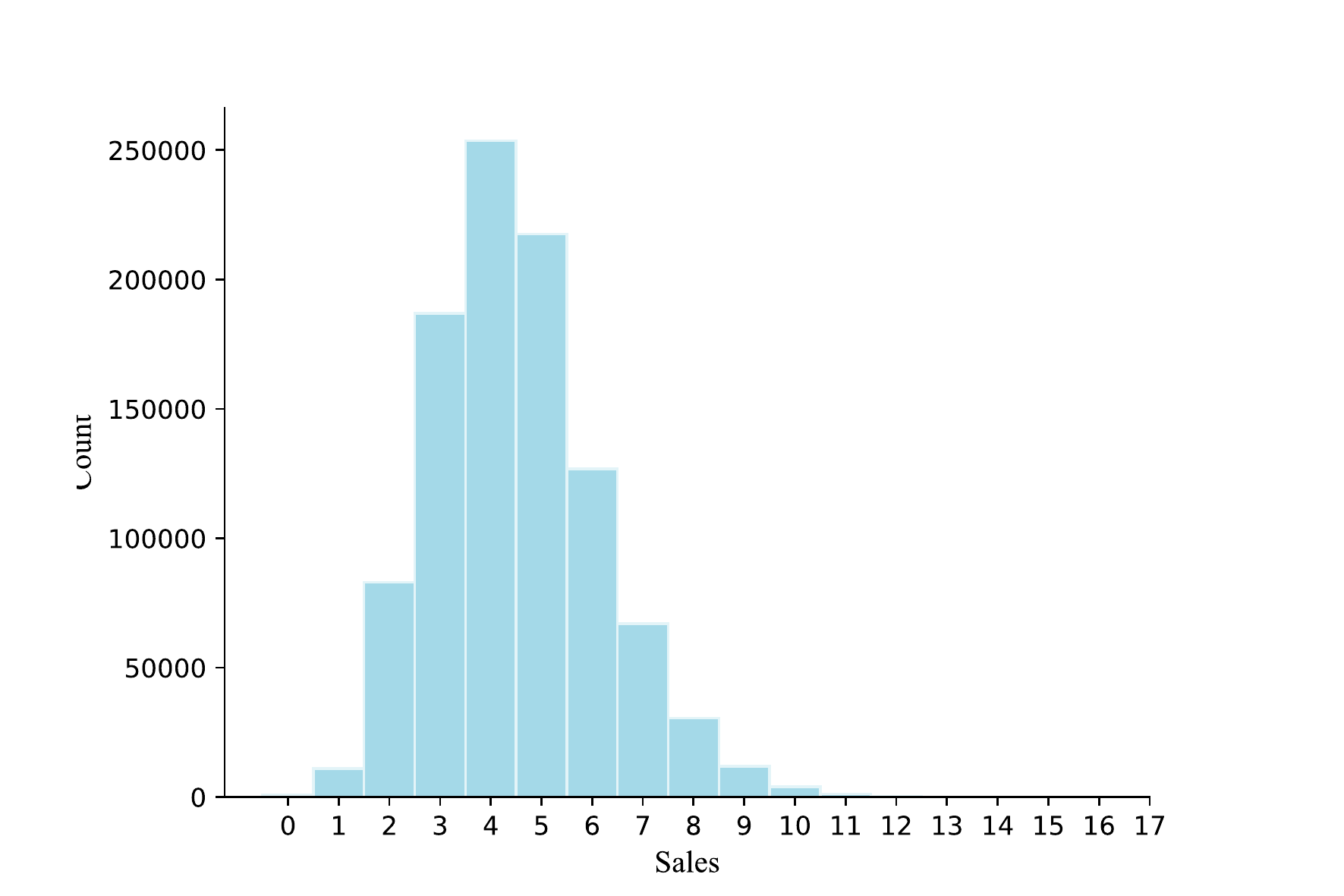}
}
\subfigure[Top 15 category of \textit{TMS} dataset.]{
\centering
\includegraphics[scale=0.25]{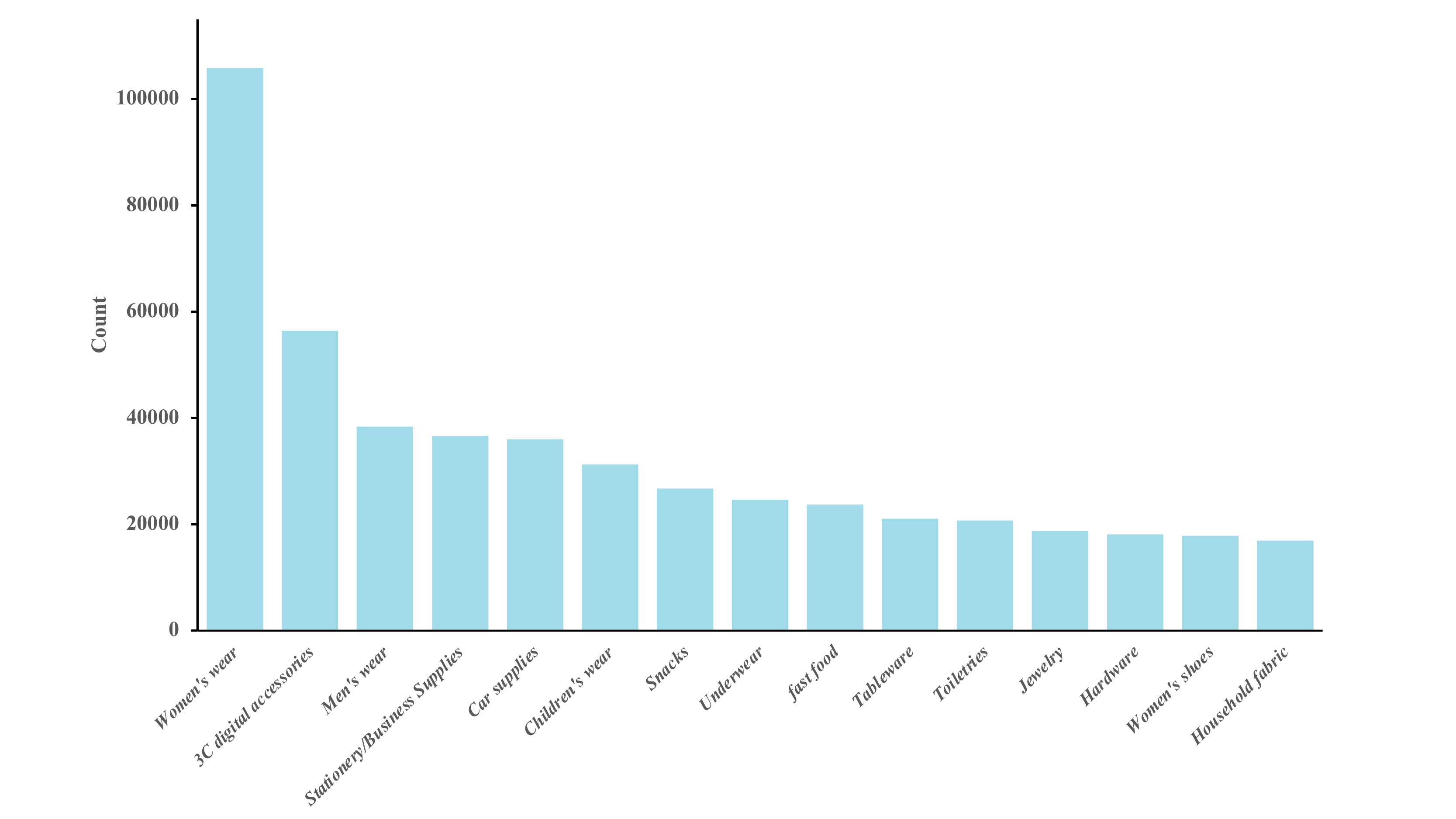}
}%
\vspace{-5pt}
\caption{Data distribution of \textit{TMS} dataset.}
\label{fig:data_dis}
\end{figure}
\vspace{5pt}

\subsection{Tmall Merchandise Sales Dataset}
The \textit{TMS} dataset is a vast real-world dataset of product sales collected from the Tmall platform, the B2C e-commerce platform of Alibaba.
To ensure its validity, we sample 1.2 million products out of billions of products by filtering out long-tail products (products with low exposure) based on preset thresholds on historical click number and sales volume.
The training, validation, and testing set are randomly divided into 80/20/20 based on its item id.
For each product, we collect its time-series data from Sep 13, 2020, to Apr 15, 2021 (215 days). The period is over half a year, during which dozens of promotion activities are held. In the experiment, we utilize the past 200 days' information and the future known knowledge to predict the following 15 days' sales volume. The products in \textit{TMS} dataset can be divided into 107 categories and 149,582 brands, and the sales value ranges from 0 to 17 after function $log\_1p$. We present the top 15 categories and sale distribution in Figure~\ref{fig:data_dis}.

\begin{figure*}[ht!]
\centering
\includegraphics[scale=0.29]{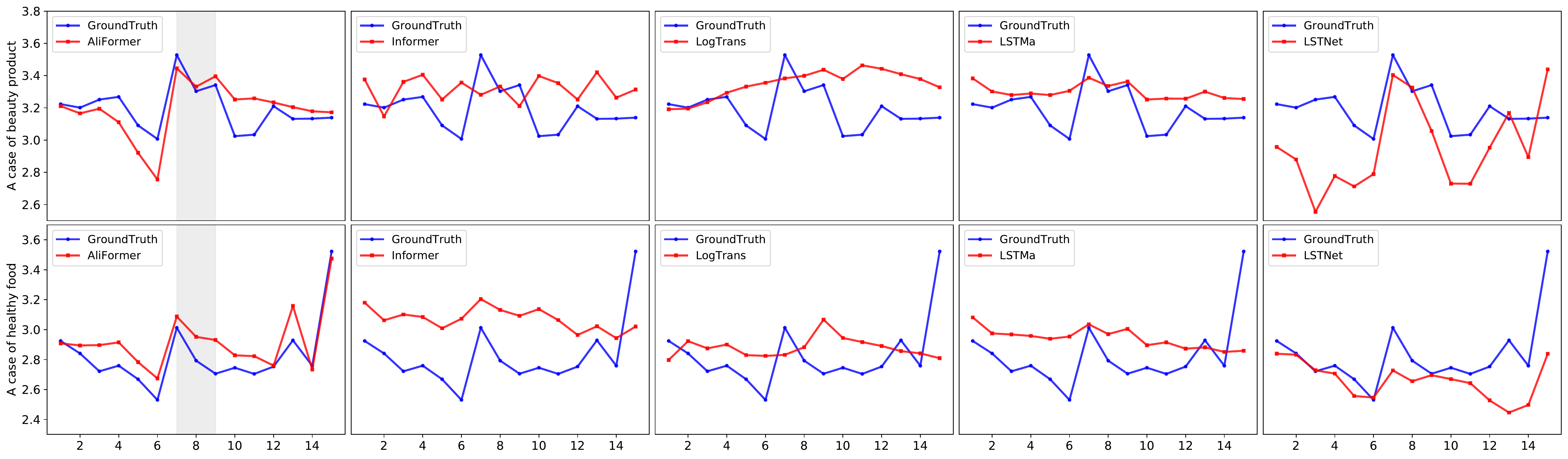}
\vspace{-10pt}
\caption{The predictions of Aliformer, Informer, LogTrans, LSTMa and LSTNet on the \textit{TMS} dataset. The red/blue curves stand for sequences of the prediction/ground truth.}
\label{fig:methods_campare}
\end{figure*}

Each sample contains 86-dimensional features, which might be divided into three groups:
\begin{itemize}
    \item ID: The primary key of each sample.
    \item Sparse Features: A feature set contains nine dims, including category, brand, activity level, date context, etc.
    \item Dense Features: There are 76 dims, including products' property and statistical information of product, seller, and category. 
\end{itemize}
Details of the 86-dimensional features are shown in Table~\ref{86dim_features}.
As we can see, the scale of \textit{TMS} (Tmall Merchandise Sales) is much larger and more complex than the current standard public time series datasets. In the actual scenario of sales forecast, we could get richer future information, such as product prices, marketing activities status in \textit{TMS}. More inspiring studies are expected on \textit{TMS}.

\section{Experimental Details}
\subsection{Baselines} We select several recently time series forecasting methods as our baselines: (1) LSTNet \cite{lai2018modeling}, which introduces the CNN with a recurrent-skip structure to extract both long- and short-term temporal patterns; (2) LSTMa \cite{bahdanau2015neural}, which introduces the addictive attention into the encoder-decoder architecture; (3) LogTrans \cite{li2019enhancing}, a variation of Transformer based on causal convolutions and the LogSparse attention; (4) Informer \cite{zhou2021informer}, the latest state-of-the-art Transformer-based model using the ProbSparse self-attention and a generative style decoder.

\subsection{Evaluation Metrics} In experiments, two metrics are used for performance evaluation: Mean Squared Error (MSE) and Mean Absolute Error (MAE). These two metrics are computed on each prediction window with stride = 1 for the whole set.

\subsection{Implementation Setting}
The five targets in experiments are listed in the following:
\begin{itemize}

    \item[-] ipv\_1d (\textit{\textbf{ipv}})
    
    \item[-] ipv\_uv\_1d (\textit{\textbf{ipv\_uv}})
    
    \item[-] pay\_ord\_amt\_1d (\textit{\textbf{gmv}})
    
    \item[-] pay\_ord\_cnt\_1d (\textit{\textbf{ord}})
    
    \item[-] pay\_ord\_byr\_cnt\_1d (\textit{\textbf{byr}})
    
\end{itemize}
We conduct grid search to tune the hyper-parameters with some ranges over the validation set. We set the prediction length $L$ progressively, i.e., $\left\{ 48, 168, 336, 720 \right\}$ in \textit{ETTh} and \textit{ECL}, $\left\{ 48, 96, 288, 672 \right\}$ in \textit{ETTm}, $\left\{ 28 \right\}$ in \textit{Kaggle-M5} and $\left\{ 15 \right\}$ in \textit{TMS}. The number of AliAttention layers used in the proposed model is selected from $\left\{ 3, 4, 5, 6, 8, 10, 12, 14 \right\}$. Each AliAttention layer consists of a 12-heads attention block. Our proposed method is optimized with the Adam \cite{kingma2015adam} optimizer and its initial learning rate of $2e^{-4}$. The training process is stopped after 20 epochs. The comparison methods are employed as recommended, and the batch size is 512. All experiments are repeated five times, and we report the averaged results. All the models are implemented in PyTorch \cite{paszke2019pytorch} and trained/tested on 8 Nvidia V100 32GB GPUs.

\begin{table*}[htp!]
  \centering
  \caption{86-dimensional features of \textit{TMS} dataset. item\_id is the primary key of each sample. Sparse features are categorized into product-related and platform-related. Dense features contain product property, product statistics, seller statistics, cate statistics and seller\_cate statistics.}
    \begin{tabular}{c|c|c}
    \toprule
    \multicolumn{3}{c}{\textbf{ID}}  \\
    \midrule
    $item\_id$ & &   \\
    \midrule
    \midrule
    \multicolumn{3}{c}{\textbf{Sparse Features}}  \\
    \midrule
    
    \multicolumn{3}{c}{$Product-related$}\\
    \midrule
    $seller\_id$ & $brand\_id$ & $cate\_level1\_id$ \\ 
    $cate\_level2\_id$ & $cate\_id$  & \\
    \midrule
    \multicolumn{3}{c}{$Platform-related$}  \\
    \midrule
    $week$ & $nday$ & $in\_activity$ \\
    $activity\_level$  & & \\
    \midrule
    \midrule
    \multicolumn{3}{c}{\textbf{Dense Features}}  \\
    \midrule
    \multicolumn{3}{c}{$Product-Property$}  \\
    \midrule
    $ brand\_score $ & $  online $ & $  price $ \\ 
     $  dsr\_fw $  & $ dsr\_wl $ & $  dsr\_zl $ \\
    \midrule
    \multicolumn{3}{c}{$Product-Statistics$}  \\
    \midrule
    $pay\_ord\_amt\_1d $ & $  pay\_ord\_amt\_1w $ & $  pay\_ord\_amt\_1m $  \\ 
    $  pay\_ord\_byr\_cnt\_1d $ &  $  pay\_ord\_byr\_cnt\_1w $ & $  pay\_ord\_byr\_cnt\_1m $\\
    $  pay\_ord\_itm\_qty\_1d $ & $  pay\_ord\_itm\_qty\_1w $ & $  pay\_ord\_itm\_qty\_1m $\\
    $  pay\_ord\_cnt\_1d $ & $  pay\_ord\_cnt\_1w $ & $  pay\_ord\_cnt\_1m $ \\
    $  cart\_cnt\_1d $ & $  cart\_itm\_qty\_1d $ & $  cart\_cnt\_1w $ \\
    $  cart\_cnt\_1m $ & $  ipv\_1d $ & $  ipv\_uv\_1d $ \\ 
    $  ipv\_1w $ & $  ipv\_uv\_1w $ & $  ipv\_1m $ \\
    $  ipv\_uv\_1m $ &  & \\ 
    \midrule
    \multicolumn{3}{c}{$Seller-Statistics$}  \\
    \midrule
    $  slr\_pay\_ord\_amt\_1d $ & $  slr\_pay\_ord\_amt\_1w $ & $  slr\_pay\_ord\_amt\_1m $ \\
    $  slr\_pay\_ord\_byr\_cnt\_1d $ & $  slr\_pay\_ord\_byr\_cnt\_1w $ & $  slr\_pay\_ord\_byr\_cnt\_1m $\\
    $  slr\_pay\_ord\_itm\_qty\_1d $ & $  slr\_pay\_ord\_itm\_qty\_1w $ & $  slr\_pay\_ord\_itm\_qty\_1m $\\
    $  slr\_pay\_ord\_cnt\_1d $ & $  slr\_pay\_ord\_cnt\_1w $ & $  slr\_pay\_ord\_cnt\_1m $\\
    $  slr\_cart\_cnt\_1d $ & $  slr\_cart\_cnt\_1w $ & $  slr\_cart\_cnt\_1m $ \\
    $  slr\_ipv\_1d $ & $  slr\_ipv\_uv\_1d $ & $  slr\_ipv\_1w $ \\
    $  slr\_ipv\_uv\_1w $ & $  slr\_ipv\_1m $ & $  slr\_ipv\_uv\_1m $\\
    \midrule
    \multicolumn{3}{c}{$Cate-Statistics$}  \\
    \midrule
    $  cate\_pay\_ord\_amt\_1d $ & $  cate\_pay\_ord\_amt\_1w $ & $  cate\_pay\_ord\_amt\_1m $ \\
    $  cate\_pay\_ord\_byr\_cnt\_1d $ & $  cate\_pay\_ord\_byr\_cnt\_1w $ & $  cate\_pay\_ord\_byr\_cnt\_1m $\\
    $  cate\_pay\_ord\_itm\_qty\_1d $ & $  cate\_pay\_ord\_itm\_qty\_1w $ & $  cate\_pay\_ord\_itm\_qty\_1m $\\
    $  cate\_pay\_ord\_cnt\_1d $ & $  cate\_pay\_ord\_cnt\_1w $ & $  cate\_pay\_ord\_cnt\_1m $\\
    $  cate\_ipv\_1d $ &  &  \\
    \midrule
    \multicolumn{3}{c}{$Seller-Cate-Statistics$} \\
    \midrule
    $  slr\_cate2\_pay\_ord\_amt\_1d $ & $  slr\_cate2\_pay\_ord\_amt\_1w $ & $slr\_cate2\_pay\_ord\_amt\_1m $ \\
    $slr\_cate2\_pay\_ord\_byr\_cnt\_1d $ & $  slr\_cate2\_pay\_ord\_byr\_cnt\_1w $ & $  slr\_cate2\_pay\_ord\_byr\_cnt\_1m $ \\
    $  slr\_cate2\_pay\_ord\_itm\_qty\_1d $ & $  slr\_cate2\_pay\_ord\_itm\_qty\_1w $  & $  slr\_cate2\_pay\_ord\_itm\_qty\_1m $ \\
    $  slr\_cate2\_pay\_ord\_cnt\_1d $ & $  slr\_cate2\_pay\_ord\_cnt\_1w $ & $  slr\_cate2\_pay\_ord\_cnt\_1m $  \\
    $  slr\_cate2\_ipv\_1d $ & $  slr\_cate2\_ipv\_uv\_1d $ &  \\
    \bottomrule
    \end{tabular}%
  \label{86dim_features}%
\end{table*}%

\section{More Visualization for Case Study}
Figure~\ref{fig:methods_campare} presents the predicted sales value by 5 models of two products randomly selected from \textit{TMS} dataset. Both two products participated in the same marketing campaign during the 7th-9th days of the period.
For the first product, the ground truth of sales value (blue curve) increases slightly during 1st-4th days and decreases in 5th-6th days due to activity inhibition. When the promotion activity began, it quickly reached its maximum, gradually decreased in the following days, and reached a stable point.
The second product's sales value shows a similar pattern as the first product, i.e., it gradually decreased in 1st-6th days, then burst in the 7th-9th days with some slow decrement. Its sales value suddenly burst on the 15th day, since the seller gave a relatively large discount on this product.

The comparisons among methods demonstrate the effectiveness of Aliformer not only in fitting trends but also in handling pulses.
Here we analyze the reasons that other baselines failed on these two cases: For the Informer, the masked attention mechanism in its decoder hinders information after the prediction moment, though it can capture rapidly changing pattern of sequence based on its feature extraction ability on historical information; LogTrans and LSTMa perform generative-style prediction, which is time-consuming and lacks the guidance of future knowledge. The Figure~\ref{fig:fig2} shows that LogTrans is effective in the short-term (1st-4th days) but accumulates error as the time interval increases; 
LSTMa tends to learn long-term trends and predicts smoother and thus can't be adapted to the fluctuations, especially during promotion activities; 
LSTNet is amenable for mining periodicity of the sequence, which can't recognize the complex patterns. Therefore, its predicted trend is contrary to the facts during the 1st-4th days in case 1 and does not fit well for case 2.

\newpage

\bibliographystyle{IEEEtran}

\begin{thebibliography}{26}
\providecommand{\natexlab}[1]{#1}

\bibitem[{Alhnaity and Abbod(2020)}]{alhnaity2020new}
Alhnaity, B.; and Abbod, M. 2020.
\newblock A new hybrid financial time series prediction model.
\newblock \emph{Engineering Applications of Artificial Intelligence}.

\bibitem[{Bahdanau, Cho, and Bengio(2015)}]{bahdanau2015neural}
Bahdanau, D.; Cho, K.; and Bengio, Y. 2015.
\newblock Neural machine translation by jointly learning to align and
  translate.
\newblock \emph{ICLR}.

\bibitem[{Bai, Kolter, and Koltun(2018)}]{bai2018empirical}
Bai, S.; Kolter, J.~Z.; and Koltun, V. 2018.
\newblock An empirical evaluation of generic convolutional and recurrent
  networks for sequence modeling.
\newblock \emph{arXiv preprint arXiv:1803.01271}.

\bibitem[{Box et~al.(2015)Box, Jenkins, Reinsel, and Ljung}]{box2015time}
Box, G.~E.; Jenkins, G.~M.; Reinsel, G.~C.; and Ljung, G.~M. 2015.
\newblock \emph{Time series analysis: forecasting and control}.

\bibitem[{Brockwell and Davis(2009)}]{brockwell2009time}
Brockwell, P.~J.; and Davis, R.~A. 2009.
\newblock \emph{Time series: theory and methods}.

\bibitem[{Chang et~al.(2018)Chang, Sun, Wu, and Lin}]{chang2018memory}
Chang, Y.-Y.; Sun, F.-Y.; Wu, Y.-H.; and Lin, S.-D. 2018.
\newblock A memory-network based solution for multivariate time-series
  forecasting.
\newblock \emph{arXiv preprint arXiv:1809.02105}.

\bibitem[{Devlin et~al.(2018)Devlin, Chang, Lee, and
  Toutanova}]{devlin2018bert}
Devlin, J.; Chang, M.-W.; Lee, K.; and Toutanova, K. 2018.
\newblock Bert: Pre-training of deep bidirectional transformers for language
  understanding.
\newblock \emph{arXiv preprint arXiv:1810.04805}.

\bibitem[{Ekambaram et~al.(2020)Ekambaram, Manglik, Mukherjee, Sajja, Dwivedi,
  and Raykar}]{ekambaram2020attention}
Ekambaram, V.; Manglik, K.; Mukherjee, S.; Sajja, S. S.~K.; Dwivedi, S.; and
  Raykar, V. 2020.
\newblock Attention based Multi-Modal New Product Sales Time-series
  Forecasting.
\newblock \emph{KDD}.

\bibitem[{Huang et~al.(2019)Huang, Wang, Wu, and Tang}]{huang2019dsanet}
Huang, S.; Wang, D.; Wu, X.; and Tang, A. 2019.
\newblock {DSAN}et: Dual self-attention network for multivariate time series
  forecasting.
\newblock \emph{CIKM}.

\bibitem[{Joshi et~al.(2020)Joshi, Chen, Liu, Weld, Zettlemoyer, and
  Levy}]{joshi2020spanbert}
Joshi, M.; Chen, D.; Liu, Y.; Weld, D.~S.; Zettlemoyer, L.; and Levy, O. 2020.
\newblock Spanbert: Improving pre-training by representing and predicting
  spans.
\newblock \emph{TACL}.

\bibitem[{Karevan and Suykens(2020)}]{karevan2020transductive}
Karevan, Z.; and Suykens, J.~A. 2020.
\newblock Transductive LSTM for time-series prediction: An application to
  weather forecasting.
\newblock \emph{Neural Networks}.

\bibitem[{Lai et~al.(2018)Lai, Chang, Yang, and Liu}]{lai2018modeling}
Lai, G.; Chang, W.-C.; Yang, Y.; and Liu, H. 2018.
\newblock Modeling long-and short-term temporal patterns with deep neural
  networks.
\newblock \emph{SIGIR}.

\bibitem[{Li et~al.(2019)Li, Jin, Xuan, Zhou, Chen, Wang, and
  Yan}]{li2019enhancing}
Li, S.; Jin, X.; Xuan, Y.; Zhou, X.; Chen, W.; Wang, Y.-X.; and Yan, X. 2019.
\newblock Enhancing the locality and breaking the memory bottleneck of
  transformer on time series forecasting.
\newblock \emph{NeurIPS}.

\bibitem[{Parmar et~al.(2018)Parmar, Vaswani, Uszkoreit, Kaiser, Shazeer, Ku,
  and Tran}]{parmar2018image}
Parmar, N.; Vaswani, A.; Uszkoreit, J.; Kaiser, L.; Shazeer, N.; Ku, A.; and
  Tran, D. 2018.
\newblock Image transformer.
\newblock \emph{ICML}.

\bibitem[{Qi et~al.(2019)Qi, Li, Deng, Cai, Qi, and Deng}]{qi2019deep}
Qi, Y.; Li, C.; Deng, H.; Cai, M.; Qi, Y.; and Deng, Y. 2019.
\newblock A deep neural framework for sales forecasting in e-commerce.
\newblock \emph{CIKM}.

\bibitem[{Qin et~al.(2017)Qin, Song, Chen, Cheng, Jiang, and
  Cottrell}]{qin2017dual}
Qin, Y.; Song, D.; Chen, H.; Cheng, W.; Jiang, G.; and Cottrell, G.~W. 2017.
\newblock A Dual-Stage Attention-Based Recurrent Neural Network for Time Series
  Prediction.
\newblock \emph{IJCAI}.

\bibitem[{Salinas et~al.(2020)Salinas, Flunkert, Gasthaus, and
  Januschowski}]{salinas2020deepar}
Salinas, D.; Flunkert, V.; Gasthaus, J.; and Januschowski, T. 2020.
\newblock DeepAR: Probabilistic forecasting with autoregressive recurrent
  networks.
\newblock \emph{IJF}.

\bibitem[{Seeger et~al.(2017)Seeger, Rangapuram, Wang, Salinas, Gasthaus,
  Januschowski, and Flunkert}]{seeger2017approximate}
Seeger, M.; Rangapuram, S.; Wang, Y.; Salinas, D.; Gasthaus, J.; Januschowski,
  T.; and Flunkert, V. 2017.
\newblock Approximate Bayesian inference in linear state space models for
  intermittent demand forecasting at scale.
\newblock \emph{arXiv preprint arXiv:1709.07638}.

\bibitem[{Seeger, Salinas, and Flunkert(2016)}]{seeger2016bayesian}
Seeger, M.; Salinas, D.; and Flunkert, V. 2016.
\newblock Bayesian intermittent demand forecasting for large inventories.
\newblock \emph{NeurIPS}.

\bibitem[{Sen, Yu, and Dhillon(2019)}]{sen2019think}
Sen, R.; Yu, H.-F.; and Dhillon, I.~S. 2019.
\newblock Think Globally, Act Locally: A Deep Neural Network Approach to
  High-Dimensional Time Series Forecasting.
\newblock \emph{NeurIPS}.

\bibitem[{Shih, Sun, and Lee(2019)}]{shih2019temporal}
Shih, S.-Y.; Sun, F.-K.; and Lee, H.-y. 2019.
\newblock Temporal pattern attention for multivariate time series forecasting.
\newblock \emph{Machine Learning}.

\bibitem[{Taylor and Letham(2018)}]{taylor2018forecasting}
Taylor, S.~J.; and Letham, B. 2018.
\newblock Forecasting at scale.
\newblock \emph{The American Statistician}.

\bibitem[{Vaswani et~al.(2017)Vaswani, Shazeer, Parmar, Uszkoreit, Jones,
  Gomez, Kaiser, and Polosukhin}]{vaswani2017attention}
Vaswani, A.; Shazeer, N.; Parmar, N.; Uszkoreit, J.; Jones, L.; Gomez, A.~N.;
  Kaiser, {\L}.; and Polosukhin, I. 2017.
\newblock Attention is all you need.
\newblock \emph{NeurIPS}.

\bibitem[{Wu et~al.(2021)Wu, Xu, Wang, and Long}]{wu2021autoformer}
Wu, H.; Xu, J.; Wang, J.; and Long, M. 2021.
\newblock Autoformer: Decomposition Transformers with Auto-Correlation for
  Long-Term Series Forecasting.
\newblock \emph{arXiv preprint arXiv:2106.13008}.

\bibitem[{Xin et~al.(2019)Xin, Ester, Bu, Yao, Li, Zhou, Ye, and
  Wang}]{xin2019multi}
Xin, S.; Ester, M.; Bu, J.; Yao, C.; Li, Z.; Zhou, X.; Ye, Y.; and Wang, C.
  2019.
\newblock Multi-task based Sales Predictions for Online Promotions.
\newblock \emph{CIKM}.

\bibitem[{Zhou et~al.(2021)Zhou, Zhang, Peng, Zhang, Li, Xiong, and
  Zhang}]{zhou2021informer}
Zhou, H.; Zhang, S.; Peng, J.; Zhang, S.; Li, J.; Xiong, H.; and Zhang, W.
  2021.
\newblock Informer: Beyond efficient transformer for long sequence time-series
  forecasting.
\newblock \emph{AAAI}.

\end{thebibliography}

\end{document}